\documentclass[conference]{IEEEtran}
\IEEEoverridecommandlockouts

\usepackage[utf8]{inputenc}
\usepackage[T1]{fontenc}

\usepackage{cite}
\usepackage{amsmath,amssymb,amsfonts}
\usepackage{algorithmic}
\usepackage{graphicx}
\usepackage{textcomp}
\usepackage{xcolor}
\usepackage{soul}
\usepackage{hyperref}

\def\BibTeX{{\rm B\kern-.05em{\sc i\kern-.025em b}\kern-.08em
T\kern-.1667em\lower.7ex\hbox{E}\kern-.125emX}}

\begin{document}

\title{Optimized Modular Design and Development of a Tilt-Rotor Bicopter Drone}


\author{Saideep Verma$^{1 \dagger}$, Nimisha Tatapudi$^{2 \dagger}$, Akshay Arjun$^{2 \dagger}$, Danwada Sharanya Sukka$^{2}$,\\ Abhishek Sarkar$^{1}$ and Joyjit Mukherjee$^{2}$
\thanks{$^{1}$Authors are with the Department of Mechanical Engineering, BITS Pilani Hyderabad Campus, Hyderabad, India.
{\tt\small f20212493h@alumni.bits-pilani.ac.in, abhisheks@hyderabad.bits-pilani.ac.in}}
\thanks{$^{2}$Authors are with the Department of Electrical and Electronics Engineering, BITS Pilani Hyderabad Campus, Hyderabad, India.
        {\tt\small f20230466@hyderabad.bits-pilani.ac.in, f20220507@hyderabad.bits-pilani.ac.in,  p20230523@hyderabad.bits-pilani.ac.in, joyjit.mukherjee@hyderabad.bits-pilani.ac.in}}%
\thanks{$^{\dagger}$ These authors have contributed equally to the work.}
}

\maketitle

\begin{abstract}
Hybrid systems like tilt-rotor bicopter drones combine the beneficial characteristics of both fixed-wing and rotary-wing technology, enabling long endurance and VTOL capability. However, such drones also require an optimum design to ensure both static and dynamic stability. The modular design of a traditional bicopter is developed in this paper based on extensive analysis and in-depth structural and aerodynamic simulations. The structural analysis has been performed to ensure that the aircraft's structure withstands the stresses encountered during different flight modes. Controlling the relative positions of the Center of Gravity (CG) and Neutral Point (NP) is an essential aspect of the design, ensuring stability during hover and positive stability during forward flight. The thrust and power analyses have been conducted to assess the flight performance and endurance. After analysis, the drone has been developed, and flight tests with a basic flight controller were conducted to validate the performance metrics obtained in the simulation.
\end{abstract}


\section{Introduction}
The aerospace industry is currently paying close attention to the development of hybrid aircraft \cite{PanigrahiKrishnaThondiyath2021} due to their ability to combine the advantages of traditional rotorcraft and fixed-wing aircraft. One such innovative design is the winged bicopter \cite{Albayarak2019}, which integrates fixed wings with a bicopter configuration to achieve vertical takeoff and landing (VTOL) capabilities \cite{KamalRamirez-Serrano2018} while enhancing maneuverability and aerodynamic efficiency. This hybrid design is being developed to overcome the limitations of traditional multirotors, such as limited flight duration and payload capacity \cite{chipade2018vtol}, by leveraging the benefits of fixed-wing aircraft \cite{Nugroho2022}. According to \cite{WuShaoWu2024}, a novel tandem dual-rotor aerial–aquatic vehicle is created by combining the characteristics of underwater autonomous underwater vehicles (AUVs) with conjunction dual-rotor unmanned aerial vehicles (UAVs). VTOL UAVs have found several applications in both the military and civilian sectors. The speed and range of fixed-wing aircraft are incorporated with the vertical lift of helicopters \cite{Su2019}. This concept aims to improve the aerial-aquatic vehicle's air flight efficiency, enabling vertical takeoff and landing for fixed-point hovering, as described in \cite{Sihuan2024}.

Currently, the winged bicopter is being designed to provide stable VTOL operations, increased payload capacity, and longer flight durations compared to conventional multirotors, as explained in \cite{AsteriaVTOL2025}. To achieve optimal performance, V/STOL missions often involve multiple transition phases \cite{Naldi2011}. When compared to traditional multirotor UAVs, the addition of a lifting wing at a particular mounting angle increases lift during forward flight, lowering energy consumption and extending range \cite{xiao2021liftingwing}. Applications such as search and rescue, medical delivery, cargo transport, and surveillance can all benefit from such hybrid configurations \cite{Osman2025hybrid}. Designs for tandem bicopters have also been investigated for maritime and amphibious aerial operations. An additive manufacturing based bicopter development approach has been discussed in \cite{wan2018tiltrotorbicopter}. Dynamic modeling and advanced control design like robust backstepping control and robust adaptive control have been explored for bicopter drones \cite{RBC2021,Cardoso2020}. Design optimization and validation of a tilt-rotor drone has been presented in \cite{Gong2025}. The design paradigm of a tilt quad-rotor UAV has been discussed in \cite{hua2025tilt}, which aims to meet requirements for enhancing the effectiveness and flight stability of UAVs in challenging conditions.

The work presented in this paper aims to address the optimized design methodology for a tilt-rotor bicopter drone to attain static as well as dynamic stability, along with desired stress and power capabilities. An in-depth static analysis has been conducted to optimize the fuselage and tilting arm designs. The arms also represent a modular design approach resulting in seamless operation. Dynamic flight analysis has been conducted to identify the various components like motors, propellers, ESCs, battery, etc. with a detailed analysis of power metrics, flight time and air speed characteristics. Finally, a hardware prototype has been developed to validate the proposed design through similar studies in an open-loop control configuration with throttle commands being sent from a controller.

The rest of the manuscript is organized as follows: The design process used for the bicopter and the corresponded static analysis of the design is presented in Section~\ref{sec:bicopter_design_static_analysis}; Section~\ref{sec:dynamic_analysis} presents the dynamic flight analysis through simulations; Section~\ref{sec:results} presents the prototype development experiment results; The paper's conclusion and possible future research directions are covered in Section~\ref{sec:conclusion}.

\section{Bicopter Design and Static Analysis} \label{sec:bicopter_design_static_analysis}
This section provides a design overview of the bicopter drone and the corresponding static analysis, both of which were crucial to optimizing the proposed design.

\subsection{Design Overview}
\begin{figure}[htp]
    \centering
    \includegraphics[width=0.85\linewidth]{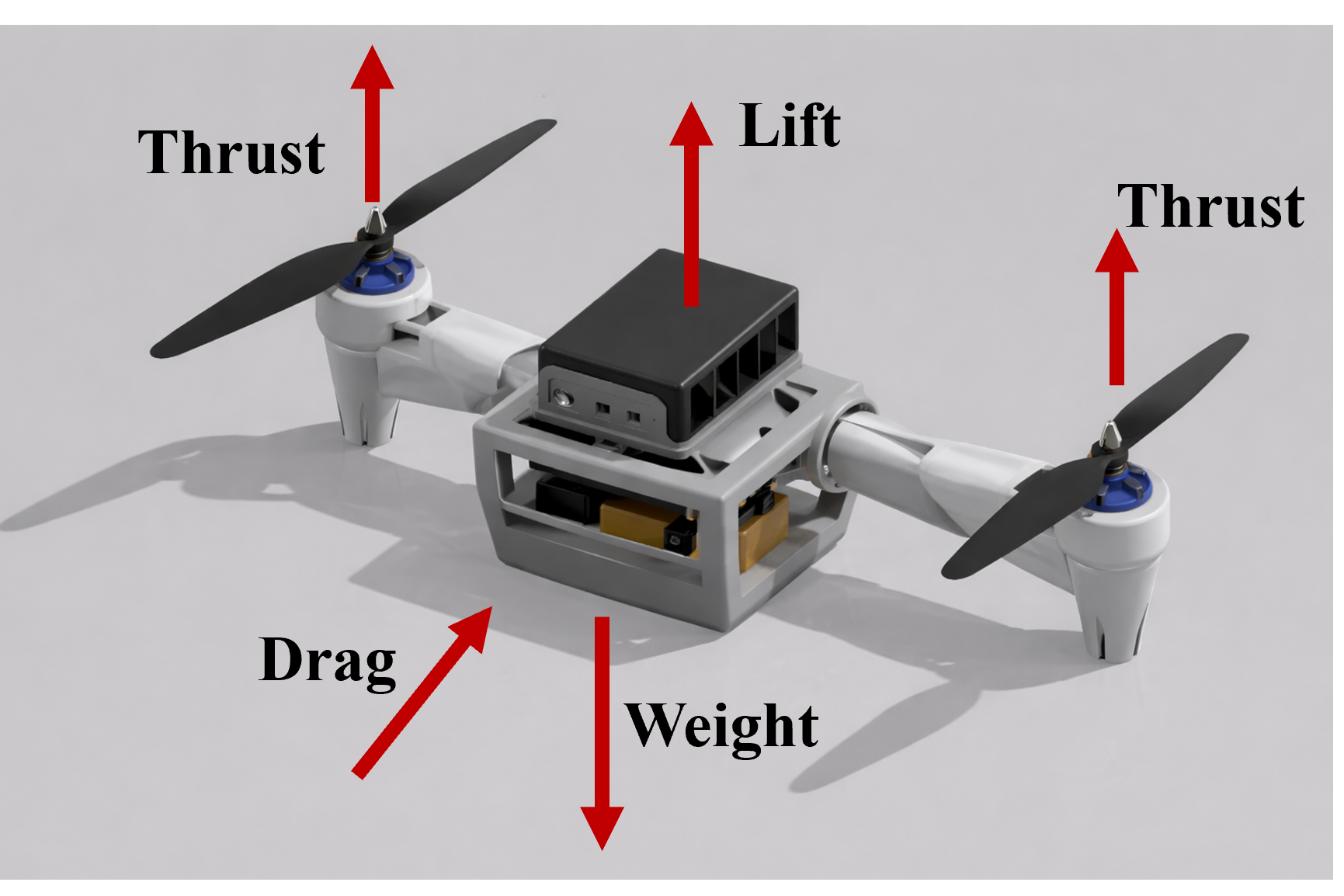}
    \caption{CAD Model of the Bicopter}
    \label{image_FBD}
\end{figure}

The bicopter’s frame is designed with modularity in mind. It has a compact rectangular fuselage that houses the battery, flight controller (FC), and ESCs in a well-organized and accessible manner, as shown in Fig. \ref{image_FBD}. The fuselage comprises compartments for the servo motors and ESCs, as well as a central battery bay, ensuring that the centre of mass is as close to the base as possible. This is a critical design aspect that ensures stability during flight. It also includes a top mount for the FC. Together, all these design aspects promote a clean wiring layout and reduce interference or signal noise. The arms of the bicopter are modular for easy replacement and are positioned to balance torque and improve stability while maintaining streamline airflow. The major physical specifications of the design are: Total Mass - $1~Kg$; Fuselage Dimensions: $14.43 \times 12.98 \times 6.98$ $cm$ ; Arm Length - $15.4~cm$, Distance between COM and Thrust Center - 3.8 $cm$.
\subsection{Stress Analysis}
The structural analysis of the bicopter motor and arm assembly performed on ANSYS, provides details on the mechanical integrity of the components under thrust loading. The analysis includes evaluation of Von Mises stress, total displacement, and safety factor, ensuring that the bicopter remains within safe limits during flight testing.

\begin{figure}[h]
    \centering
    \includegraphics[width=0.95\linewidth]{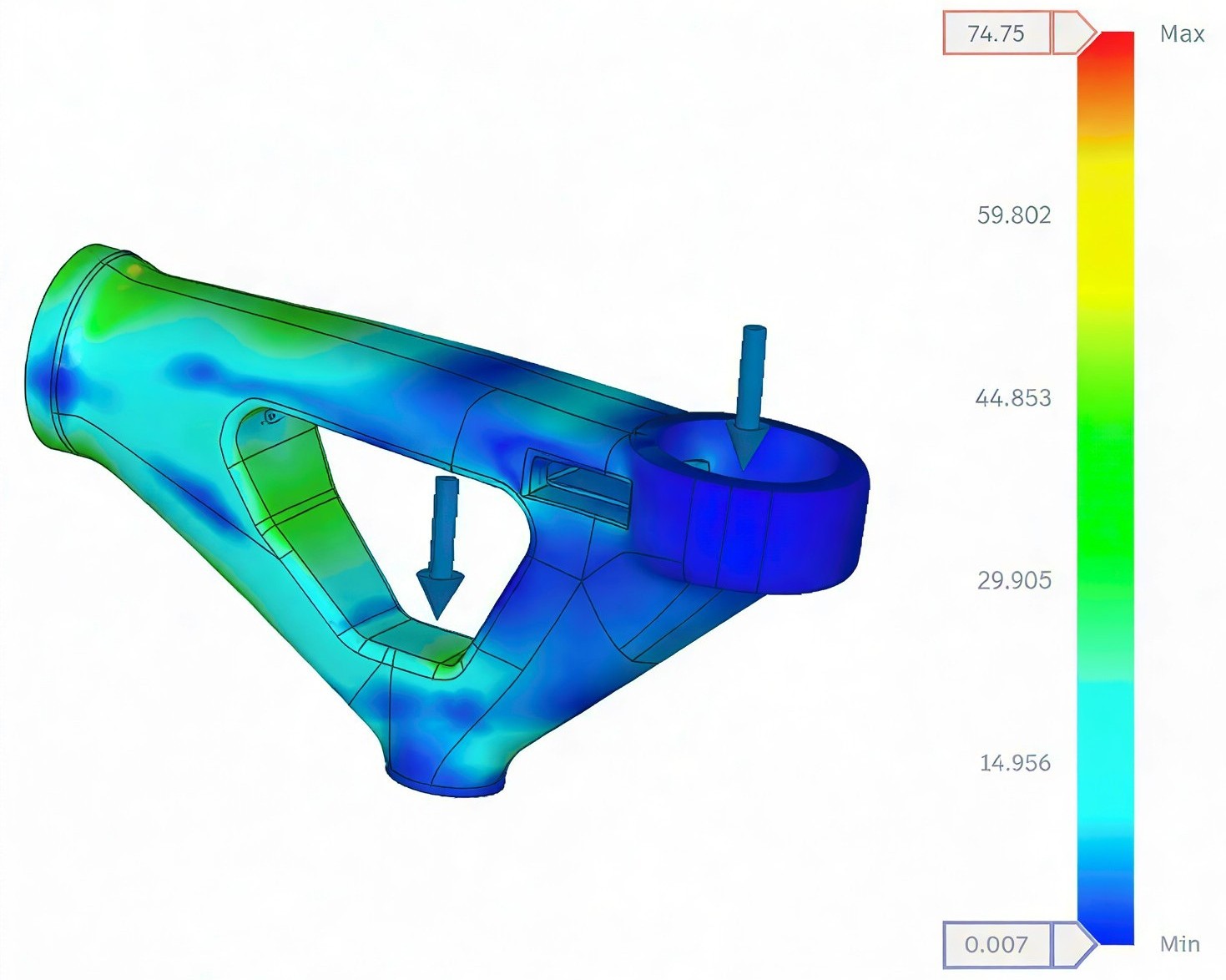}
    \caption{Stress Analysis}
    \label{image_stress analysis}
\end{figure}

Considering the mass and maximum thrust of the propellers, the drone will be exposed to a maximum load of $40-50N$, considering a $1.38$ factor of safety. The stress analysis of the arm has been performed considering much higher downforce ($100~N$)to test the component till its failure condition. The peak Von Mises stress recorded is $14.98 MPa$, localized near the junction where the motor mounts onto the arm. This is expected due to the direct load transmission from the propeller’s thrust and torque, as shown in Fig. \ref{image_stress analysis}. The selected material is PLA Aero, with a yield strength of around $24~MPa$, which meets the requirement based on the analysis. The detailed results of the stress analysis are given in Table \ref{tab:stress}.

\begin{table}[h]
\centering
\caption{Stress Analysis Results}
\label{tab:stress}
\scriptsize
\begin{tabular}{|l|c|c|}
\hline
\textbf{Name} & \textbf{Minimum} & \textbf{Maximum} \\
\hline
von Mises & 1.927e-04 MPa & 14.427 MPa \\
First Principal & -2.689 MPa & 8.516 MPa \\
Third Principal & -14.653 MPa & 2.671 MPa \\
Normal XX & -14.261 MPa & 7.333 MPa \\
Normal YY & -3.798 MPa & 5.009 MPa \\
Normal ZZ & -3.791 MPa & 3.672 MPa \\
Shear XY & -2.93 MPa & 2.168 MPa \\
Shear YZ & -1.385 MPa & 1.323 MPa \\
Shear ZX & -3.41 MPa & 3.586 MPa \\
\hline
\end{tabular}
\end{table}

\subsection{Safety Factor}
\begin{figure}[htp]
    \centering
    \includegraphics[width=0.95\linewidth]{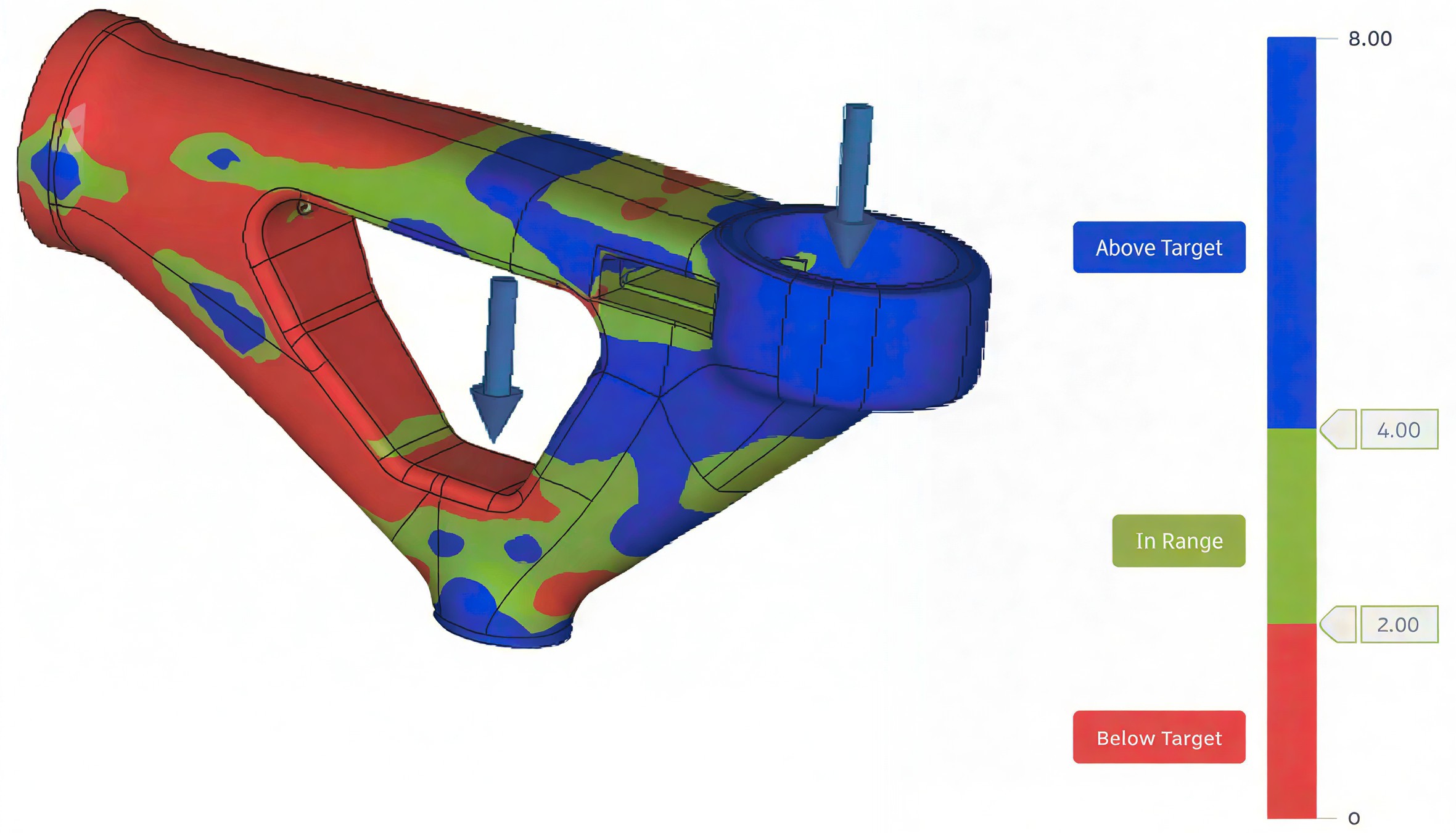}
    \caption{Safety Factor}
    \label{image_static analysis}
\end{figure}

The minimum safety factor is 1.38, occurring at the motor mount region as shown in Fig. \ref{image_static analysis}. The result suggests that the design is suitable for development and safe for flight. However, to enhance overall safety, additional reinforcement or material additions can be applied in critical stress zones to ensure the structure withstands load conditions during flight.

\begin{figure*}[h]
    \centering
    \includegraphics[width=7.2in]{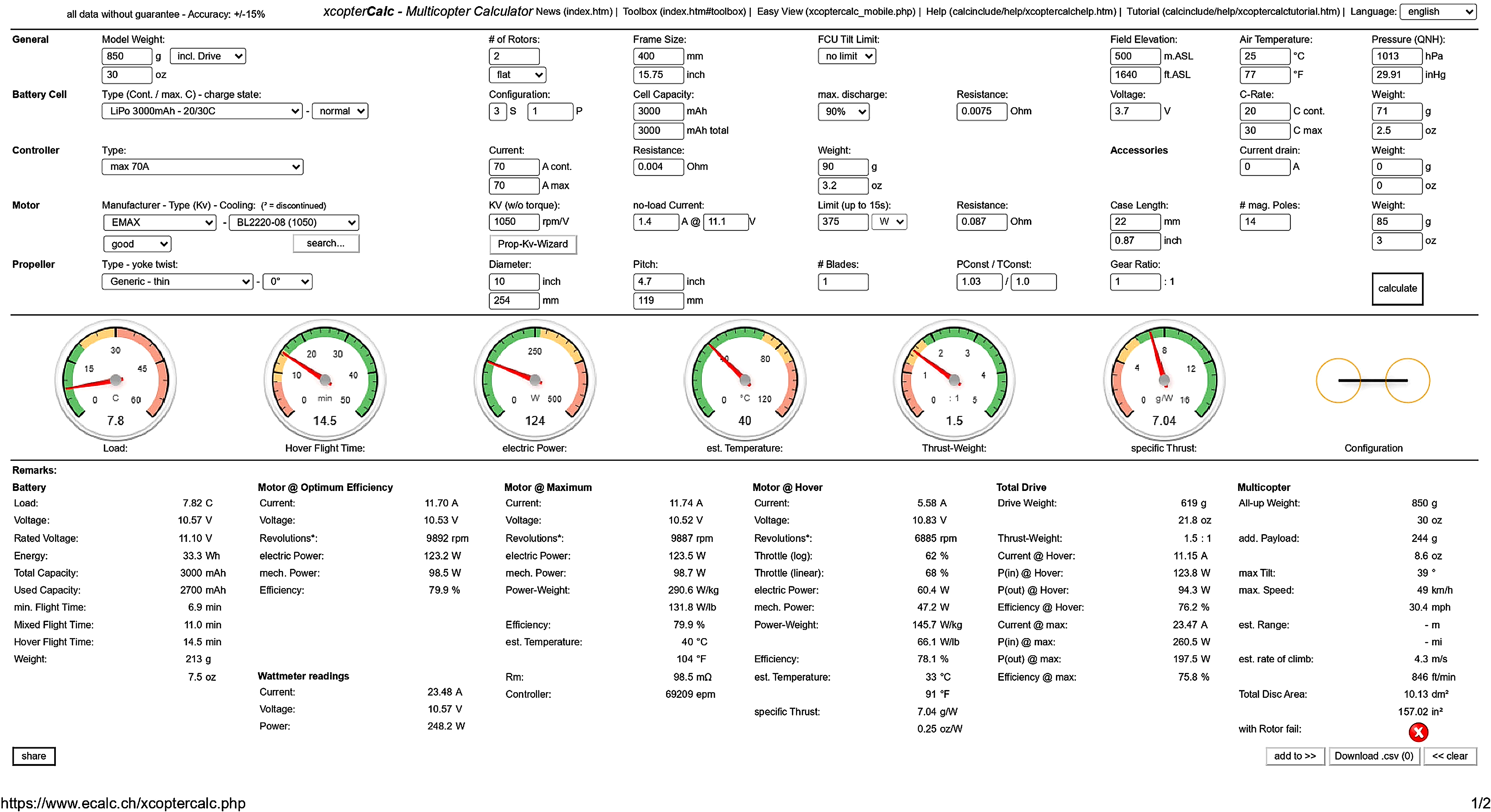}
    \caption{Flight Time and Air Speed Characteristics}
    \label{image_simulation results}
\end{figure*}

\subsection{Displacement}
The maximum displacement observed is $2.537 mm$, concentrated at the motor housing, with a gradual decrease along the arm. This deformation is within acceptable limits and will not greatly affect flight stability or propeller alignment, especially considering dynamic damping during flight. The total displacement as well as the displacement along each axis is given in Table \ref{tab:displacement}.
\begin{table}[h]
\centering
\caption{Displacement Results}
\label{tab:displacement}
\scriptsize
\begin{tabular}{|c|c|c|}
\hline
\textbf{Displacement Parameter} & \textbf{Minimum } & \textbf{Maximum } \\
\hline
\texttt{Total} & $0.00$ mm  & $1.051$ mm \\
\texttt{X} & $-0.434$ mm & $0.191$ mm  \\
\texttt{Y} & $-1.0337$ mm  & $9.598 \times 10^{-4}$ mm  \\
\texttt{Z} & $-0.012$ mm & $0.021$ mm \\
\hline
\end{tabular}
\end{table}
\section{Dynamic Flight Analysis} \label{sec:dynamic_analysis}
To evaluate the performance characteristics of the bicopter design before physical prototyping, a comprehensive analysis was conducted using eCalc xcopterCalc \cite{epropcalc_easy}, a reliable tool for multicopter propulsion analysis. The flight time and air speed characteristics for the proposed design are shown in Fig. \ref{image_simulation results}.
\subsection{Dynamic Performance Insights}
Various performance metrics under flight and dynamic conditions were analyzed to validate the design's efficacy. The aerodynamic and thrust forces, along with the weight of the drone, are shown in Fig. \ref{image_FBD}.
The motor curve during dynamic conditions illustrates the following:
\begin{itemize}
    \item A linear rise in electrical power with current;
    \item A peak in efficiency around $12A$;
    \item A relatively flat RPM curve beyond $8A$, showing consistent thrust across
throttle ranges;
    \item Gradual waste power buildup and thermal rise, but well below critical
thresholds.
\end{itemize}

This confirms that the specification of the chosen $1700KV$ motor with a $10 \times 4.7$ prop is well-matched and safely
loadable up to $~15A$ of continuous current.

\begin{figure*}[t]
    \centering
    \includegraphics[width=6in]{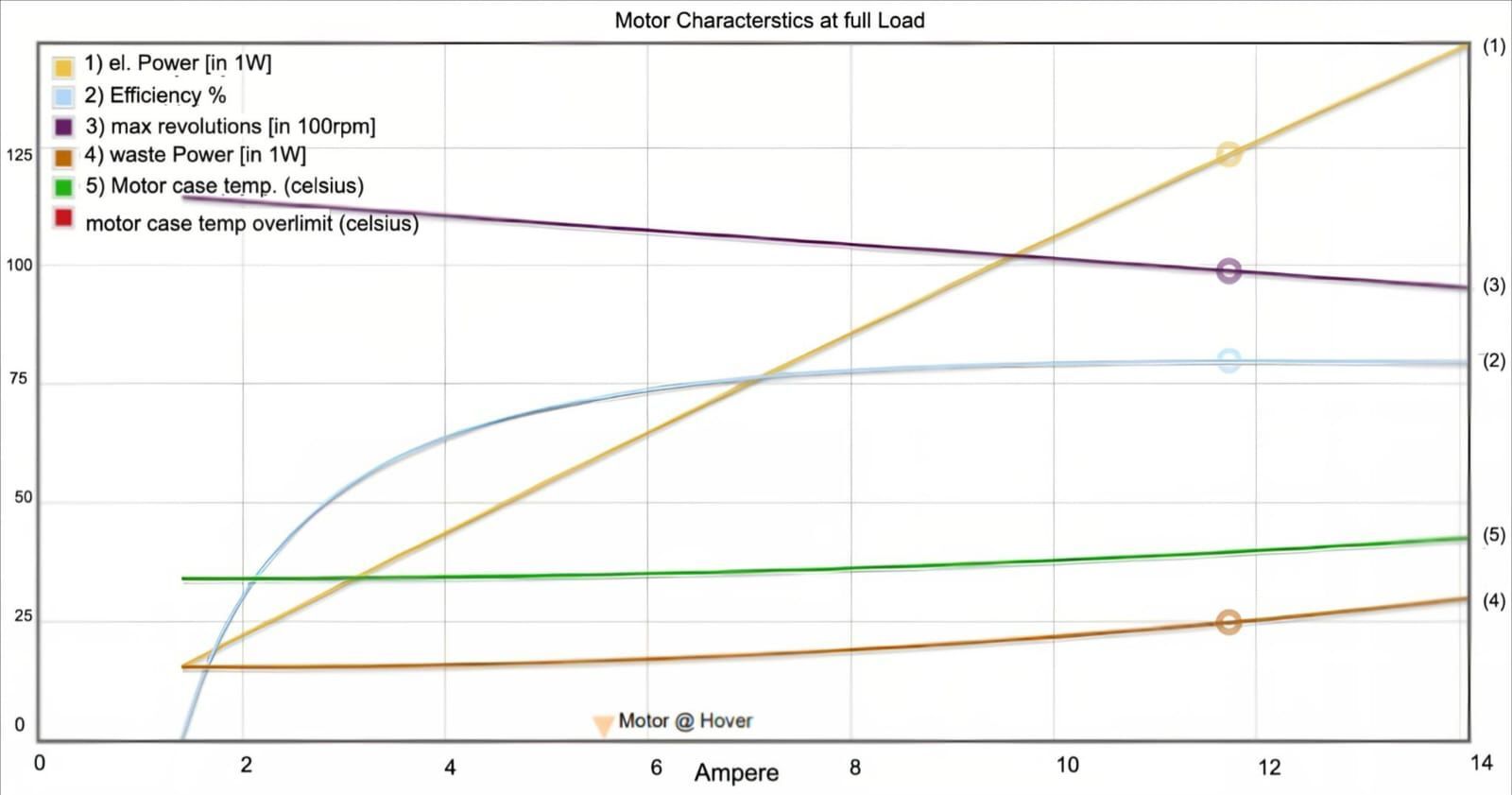}
    \caption{Motor Characteristics at full load}
    \label{image_Motor Characteristics at full load}
\end{figure*}

\begin{figure*}[t]
    \centering
    \includegraphics[width=6in]{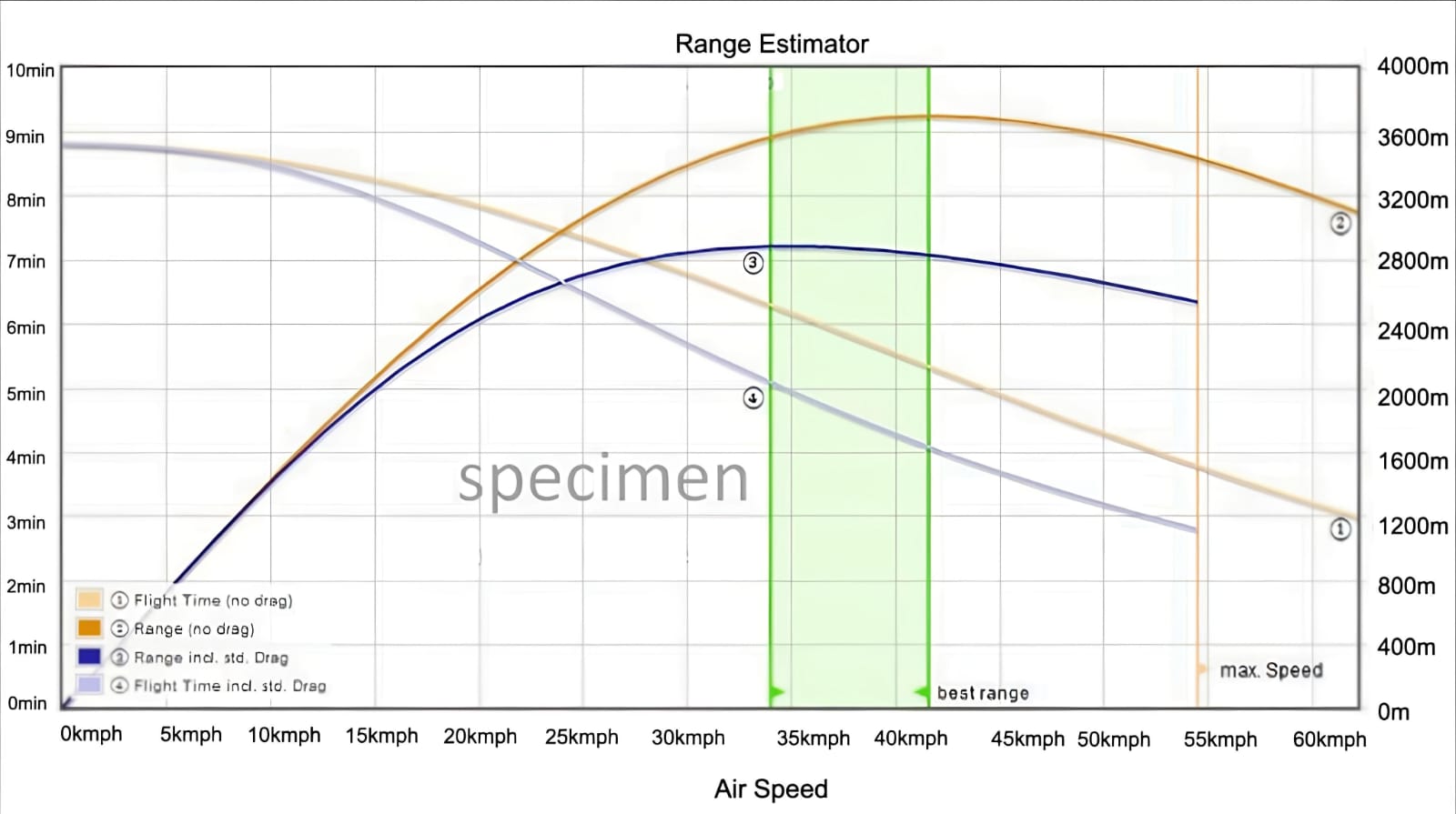}
    \caption{Range Estimator}
    \label{image_Range Estimator}
\end{figure*}

\begin{table}[h]
\centering
\caption{Analysis Results Summary}
\label{tab:results_summary}
\scriptsize
\begin{tabular}{|l|c|c|}
\hline
\textbf{Name} & \textbf{Minimum} & \textbf{Maximum} \\
\hline
Safety Factor & 1.386 & 15.00 \\
Safety Factor (Per Body) & 1.386 & 15.00 \\
\hline
\multicolumn{3}{|c|}{\textbf{Reaction Force}} \\
\hline
Total & 0.00 N & 16.235 N \\
X & -10.685 N & 15.892 N \\
Y & -2.833 N & 5.383 N \\
Z & -3.817 N & 5.229 N \\
\hline
\multicolumn{3}{|c|}{\textbf{Strain}} \\
\hline
Equivalent & 1.456e-07 & 0.007 \\
First Principal & 1.030e-07 & 0.005 \\
Third Principal & -0.008 & -1.305e-07 \\
Normal XX & -0.006 & 0.003 \\
Normal YY & -0.001 & 0.003 \\
Normal ZZ & -0.004 & 0.002 \\
Shear XY & -0.004 & 0.003 \\
Shear YZ & -0.003 & 0.003 \\
Shear ZX & -0.004 & 0.004 \\
\hline
\end{tabular}
\end{table}

\subsection{Thermal Efficiency Insights}
Motor thermal safety is essential for performance longevity. The maximum motor temperature observed is $40^0C$, and the estimated hover temperature from simulation is $33^0C$, confirming efficient cooling due to passive airflow and proper load distribution within the system. The motor efficiency at hover is approximately $78\%$, and at the optimum operating point ($12A$), it
reaches $80\%$, which is excellent for this class of motors. This demonstrates that the motor operates well within its optimal RPM and torque range. Additionally, the power-to-weight ratio at hover is about $145.7 W/kg$, providing sufficient agility while balancing energy use. Since there is no overheating or efficiency drop, the model assures long-term durability under varied flight conditions.

\subsection{Hover and Power Metrics} \label{power_metric}
The total model weight is 850 grams. Consequently, each motor must consistently produce at least 640 grams of thrust for stable hover. The simulation validates that the use of EMAX 1700 KV BLDC motors and $10" \times 4.7"$ propellers meets this requirement with a thrust to weight ratio of approximately 2:1, ensuring the capability of lift and maneuverability. While calculations suggest a maximum thrust of $31.18 N (3.17 kg)$ per motor, an operational baseline of 2:1 is widely accepted as optimal ratio \cite{Mogorosi2021} as it prioritizes thermal efficiency and provides the necessary control margin for differential thrust and quick tilt-servo stabilization.

At hover, each motor draws around 6A, and the total power required is nearly 125 watts. The motor characteristics at full load for the proposed design is shown in Fig. \ref{image_Motor Characteristics at full load}. The specific thrust is 7.5 g/W, a decent benchmark for energy efficiency in multirotor systems. At maximum throttle, it draws about 25A and consumes a peak power of 260.5W, remaining within the capabilities of the 3S LiPo battery and 30A ESCs.
The Centre Of Mass stands at a mere 35mm above the ground with 3 points of contact, namely the base of the fuselage, with area $8158.2~mm^2$ and the bases of the rotor motors with area  $123.2~mm^2$. Hence, the support to the COM is a cumulative of these surface areas, providing static stability during takeoff.

\subsection{Flight Time Estimates}
Flight endurance is a key requirement in aerial vehicles, especially for a bicopter. Using a $2200mAh$ 3S LiPo (35C) battery, the Minimum Flight Time ($100\%$ throttle) was found to be $7$ minutes, Mixed Use Flight Time (moderate throttle cycle) was found to be $11$ minutes and the Stable Hover Flight Time was obtained as $14.5$ minutes.
The battery discharge is well managed, with $1980mAh$ of usable capacity, indicating nearly full utilization without exceeding the safe discharge limits. These results validate the design for short to mid duration missions such as surveillance, inspection, and test flights with a margin for return to launch or landing. Fig. \ref{image_Range Estimator} shows the estimated range and corresponding flight time for the operation of drone at specific air speeds. 

The simulation indicates that it provides outstanding hovering stability, efficient motor loading, sufficient thrust, extended hover times, and safe thermal performance. 
\begin{figure}[htp]
    \centering
    \includegraphics[width=0.95\linewidth]{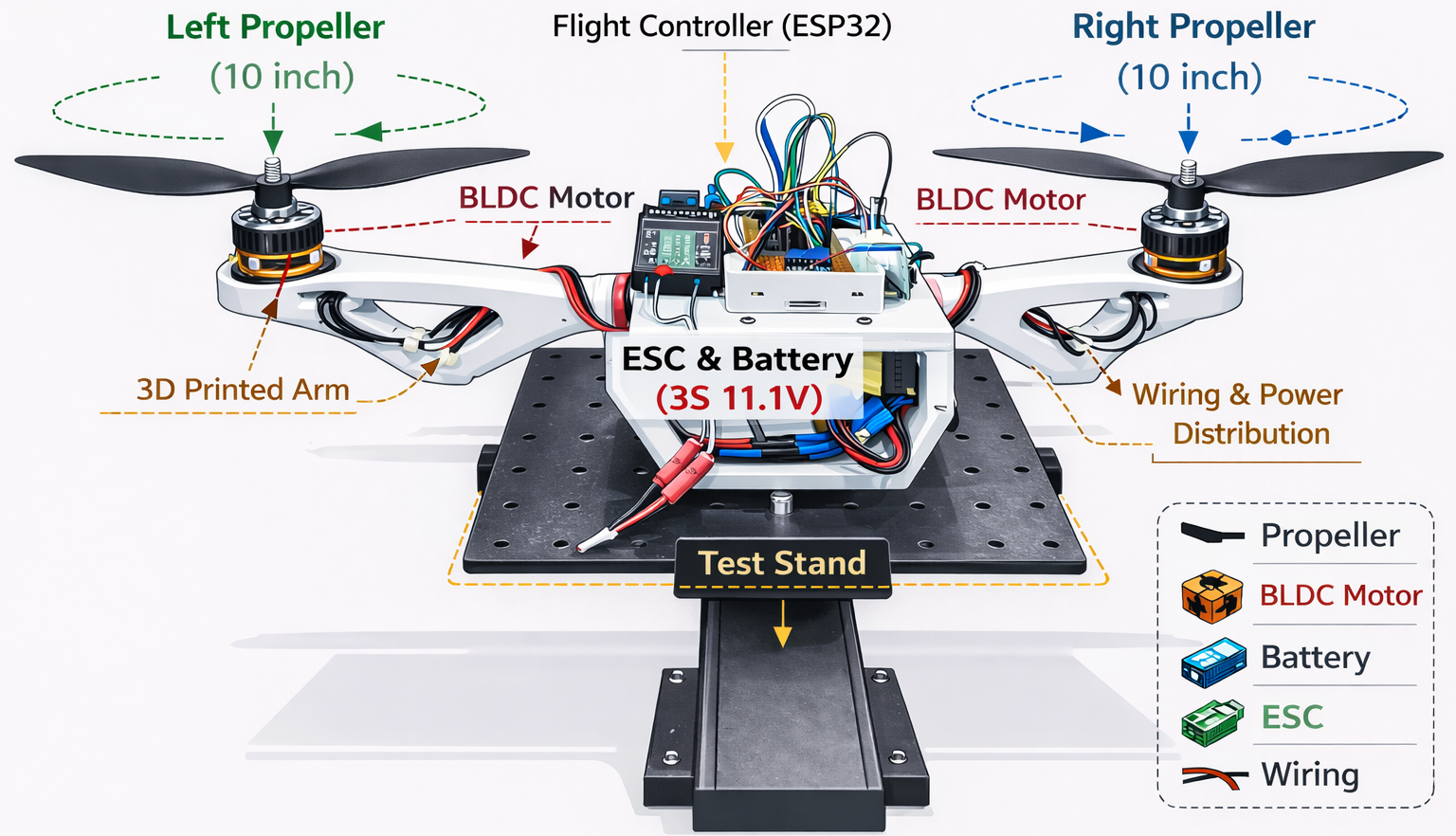}
    \caption{Hardware prototype and Experimental setup of Bicopter}
    \label{hardware_bicopter}
\end{figure}

\section{Prototyping and Experimental Results} \label{sec:results}
This section presents the hardware prototype development of the proposed bicopter design and corresponding experimental verification of the power, attitude, and actuation metrics. Basic hover and pitch-forward attitudes are tested with the bicopter on a gyroscopic Drone Testing Stand.

\begin{figure}[h]
    \centering
\includegraphics[width=0.9\linewidth]{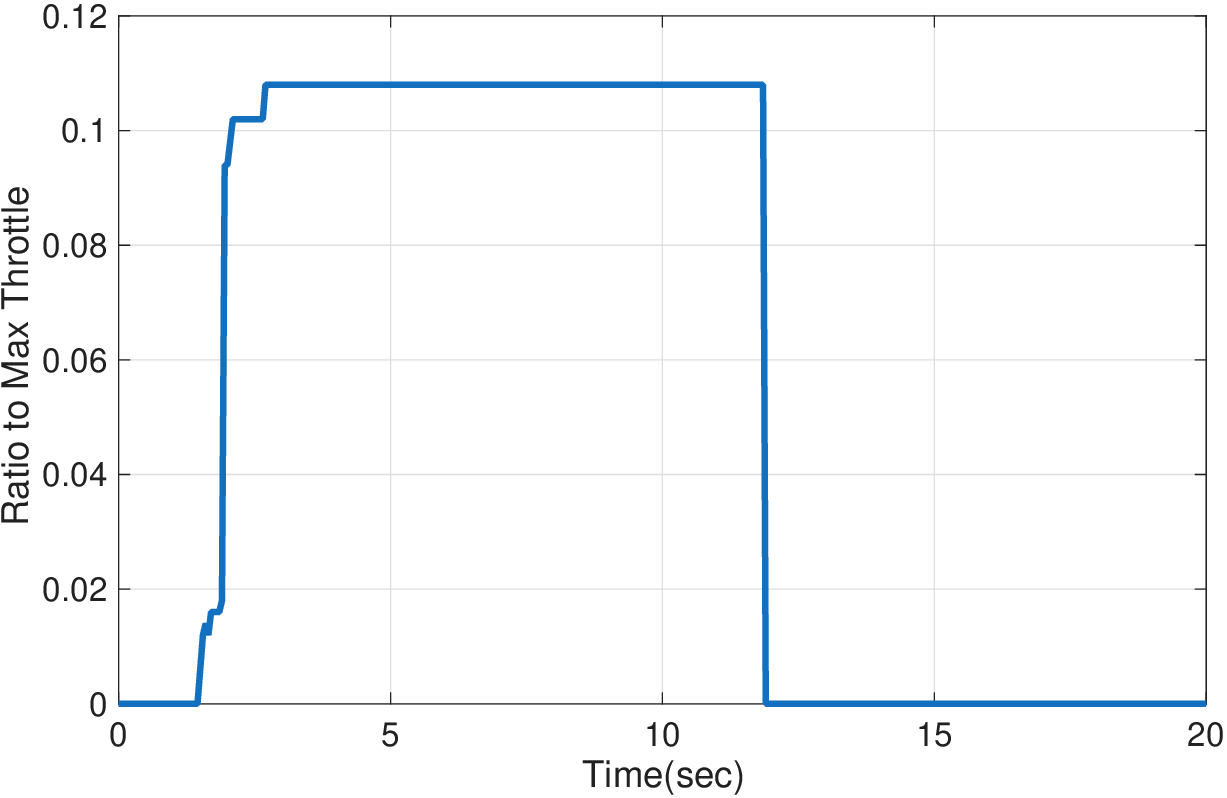}
    \caption{Throttle for Hovering Condition}
    \label{Hover_Throttle}
\end{figure}
\begin{figure}[h]
    \centering
\includegraphics[width=0.9\linewidth]{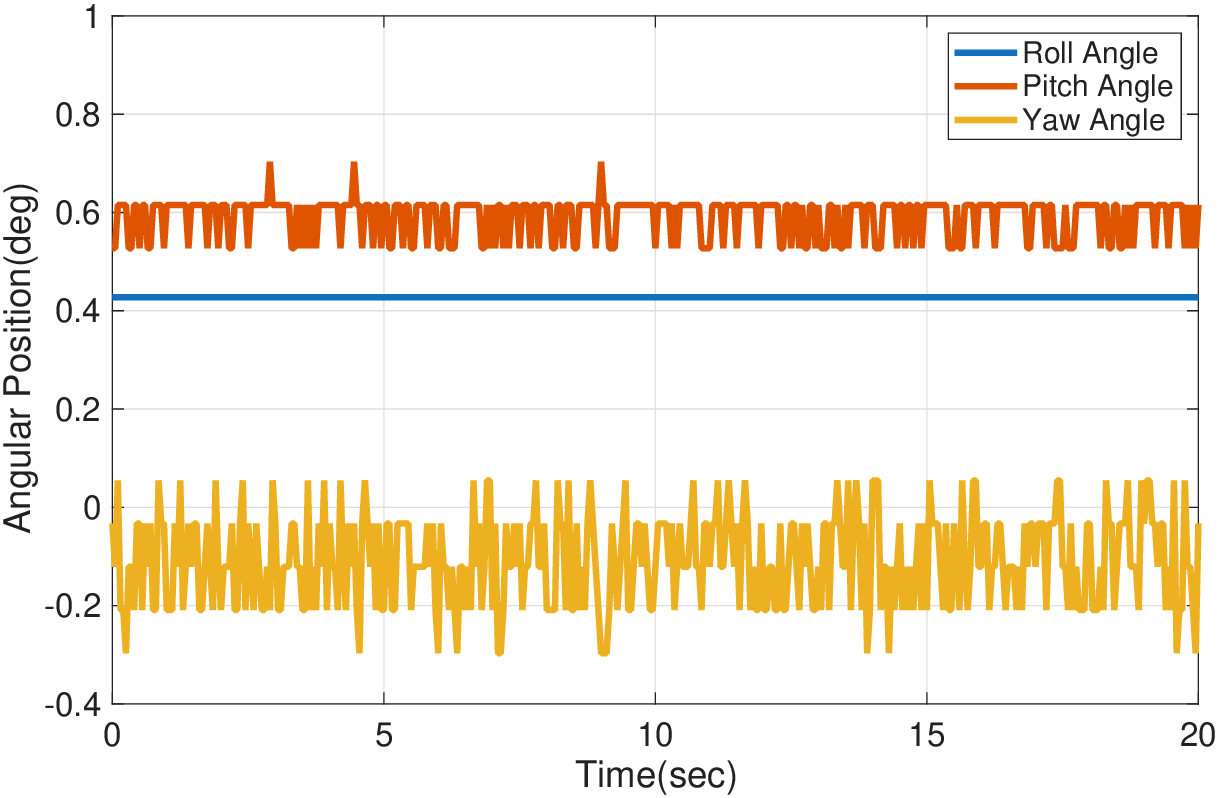}
    \caption{Drone Attitude for Hovering Condition}
    \label{Hover_Orient}
\end{figure}
\begin{figure}[h]
    \centering
\includegraphics[width=0.9\linewidth]{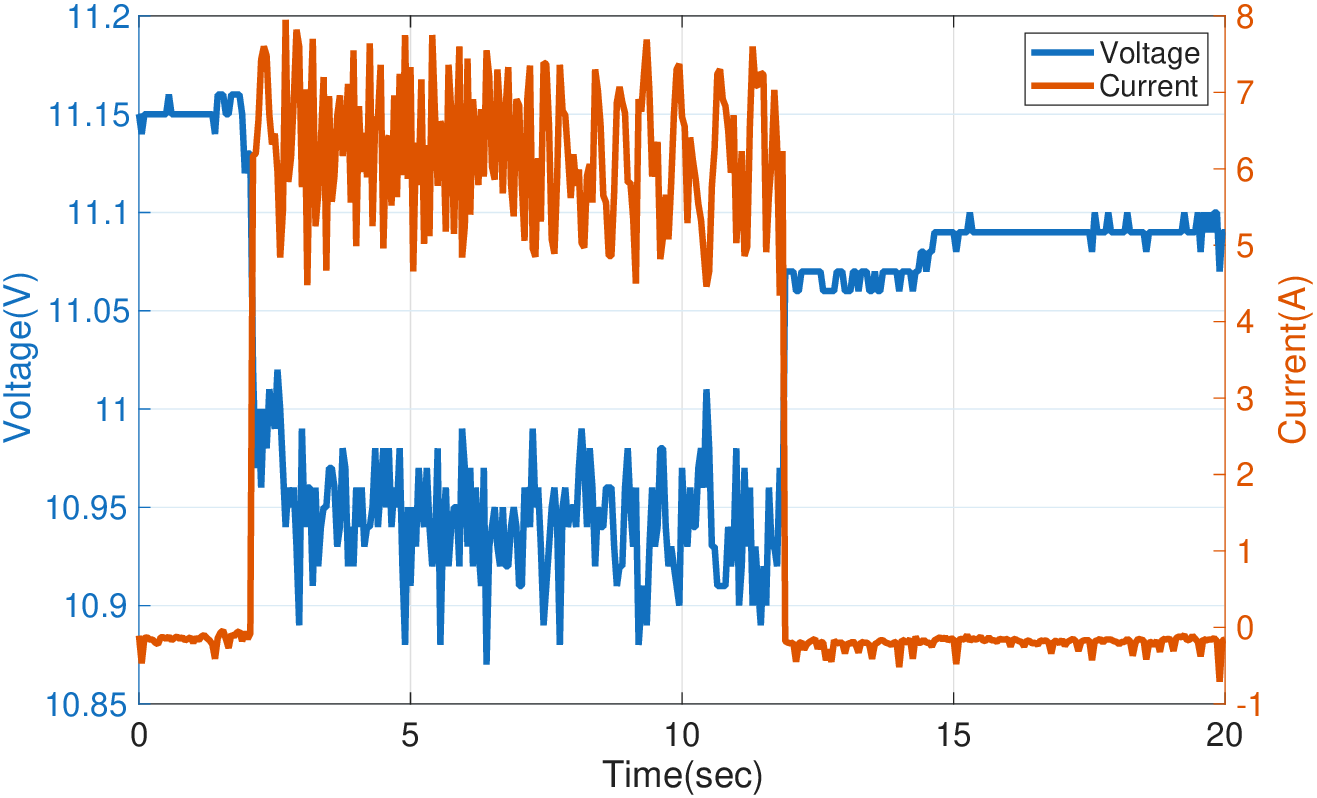} 
    \caption{Voltage and Current for Hovering Condition}
    \label{Hover_Vol_}
\end{figure}
\begin{figure}[h]
    \centering
\includegraphics[width=0.9\linewidth]{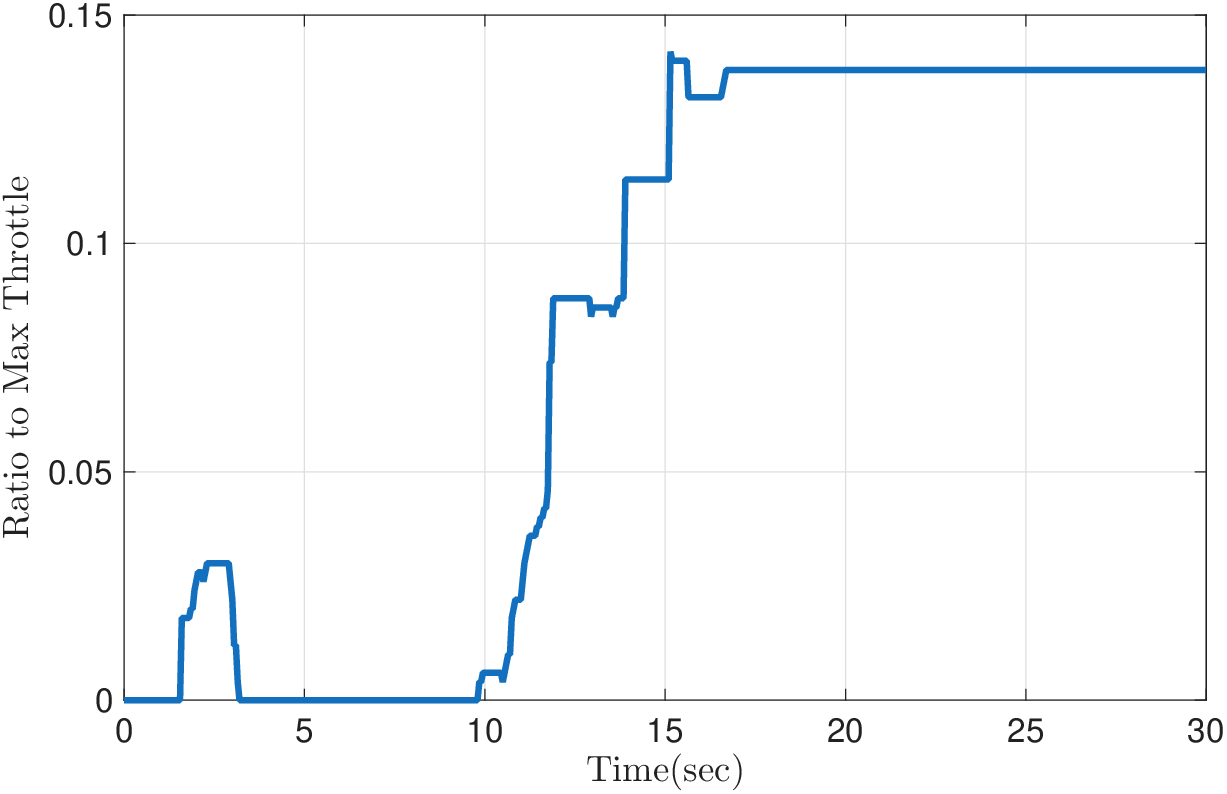}
    \caption{Throttle for Pitch Condition}
    \label{pitchthrottle}
\end{figure}
\begin{figure}[h]
    \centering
\includegraphics[width=0.9\linewidth]{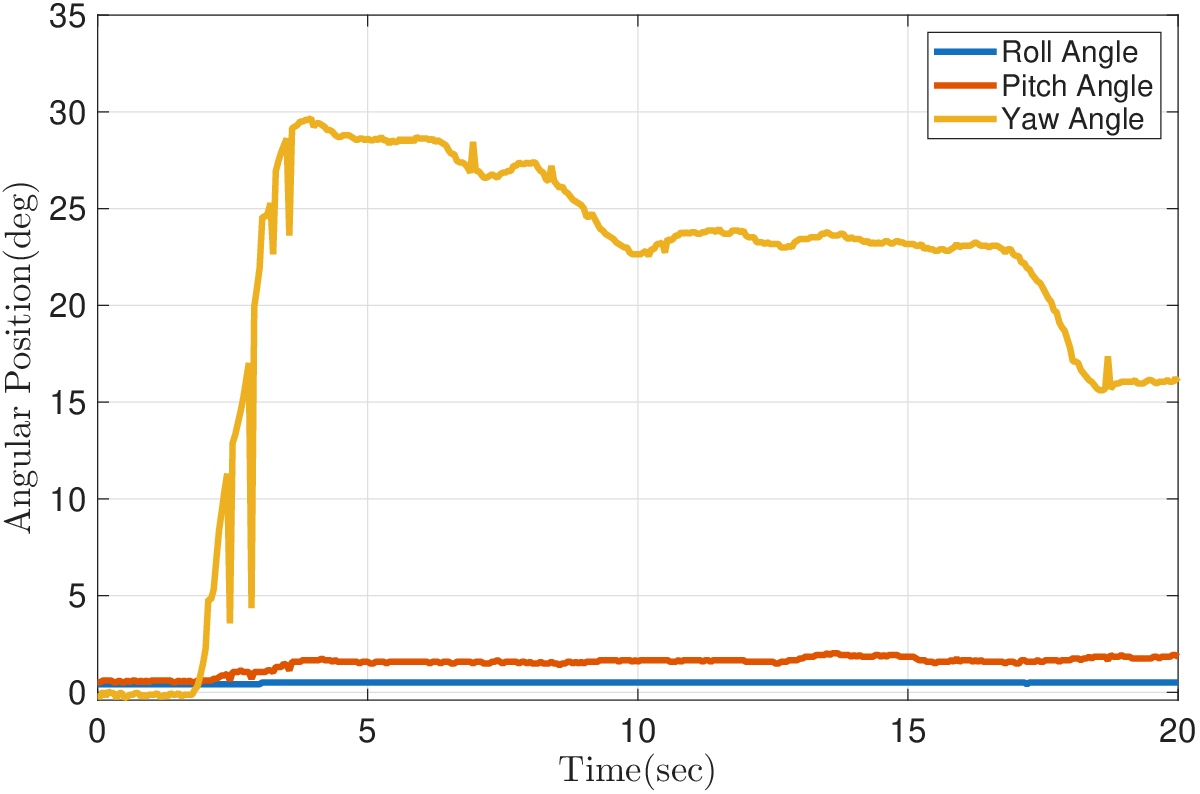}
    \caption{Drone Attitude for Pitch Condition}
    \label{throttle}
\end{figure}
\begin{figure}[h]
    \centering
\includegraphics[width=0.9\linewidth]{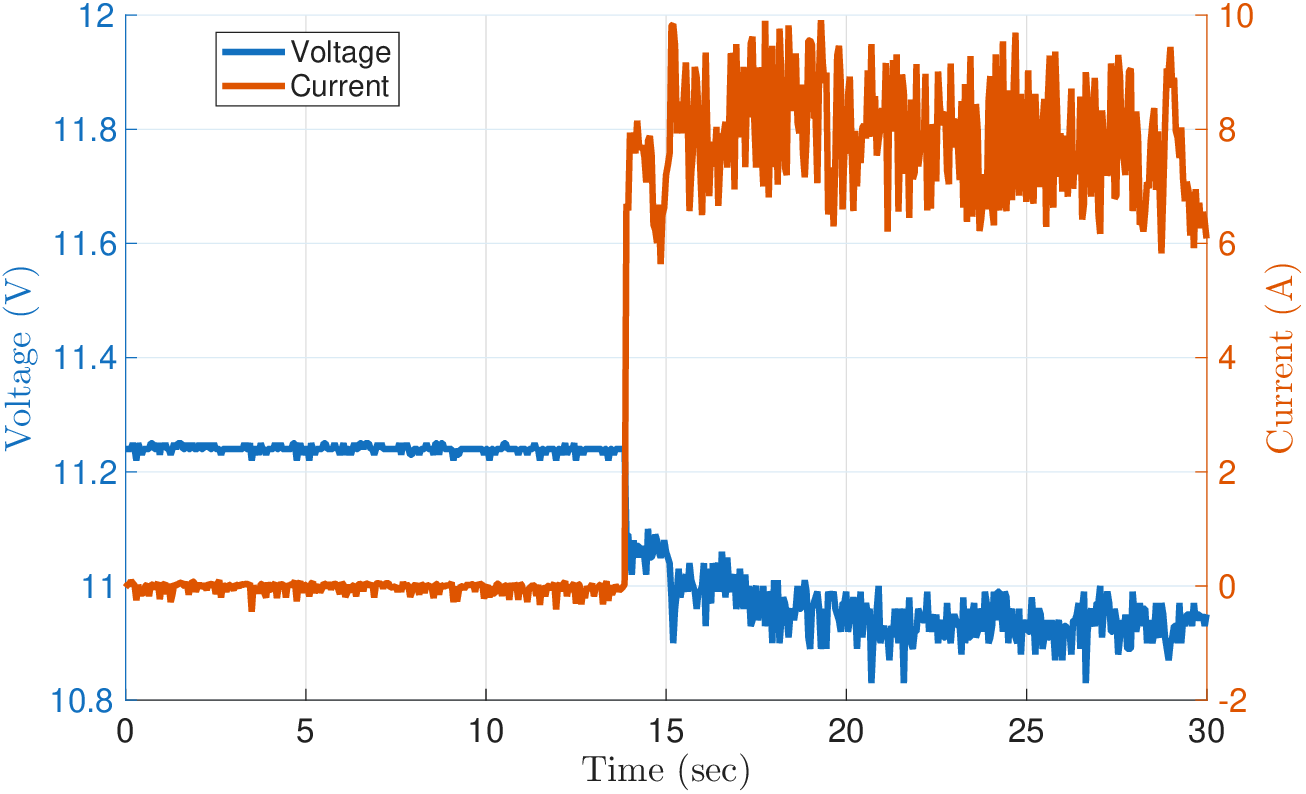} 
    \caption{Voltage and Current for Pitch Condition}
    \label{pitchvc}
\end{figure}

\subsection{Prototyping Overview}
To validate the proposed design, a hardware prototype has been developed, as shown in Fig. \ref{hardware_bicopter}. The components employed are listed in Table \ref{tab:hardware_components} along with their specifications.

Material selection plays a vital role in performance. The fuselage and arms have been 3D printed in PLA Aero, a material that foams during printing, expanding in volume to create lightweight parts with reduced material usage. It maintains good stiffness and layer bonding, making it ideal for aerospace and drone applications. FDM (Fused Deposition Modeling) 3D printing technique has been used to develop the prototype. Passive cooling is facilitated through vents in the fuselage, ensuring heat dissipation for ESCs and other electronics during extended flight times. The IMU employed for localization communicates with the Teensy Microcontroller via I2C/SPI protocol. The ESP32 communicates with Teensy via WiFi with an approximate telemetry range of $1~Km$ with an additional antenna. The experimental results have been obtained in an open-loop control scenario.

\begin{table}
\centering
\caption{Hardware Component Specifications}
\label{tab:hardware_components}
\scriptsize
\begin{tabular}{|l|c|}
\hline
\textbf{Component} & \textbf{Specification} \\
\hline
3D Printing Material & PLA Aero \\
Controller & Teensy 4.0 Microcontroller (32-bit dual-core) \\
Inertial Measurement Unit (IMU) & MPU6050 \\
Current Sensor & ACS712\\
Voltage Sensor & VCC < 25 V\\
BLDC Motors & EMAX 1700 KV BLDC\\
Propellers & 10" (2-Blade, Nylon-ABS) \\
Electronic Speed Controller & 30A SimonK ESC\\
Telemetry & ESP32 (Wifi) \\
Battery & Bonka 3S 2200 mAH LiPo battery \\
Drone Testing Stand & Eureka Dynamics FFT Gyro 4.0 (Aluminum) \\
\hline
\end{tabular}
\end{table}

\subsection{Results and Discussion}
 The primary objectives of this section are to validate the proposed design and to assess attitude stability and power consumption with respect to the throttle for hover and pitch forward conditions.  Fig.~\ref{Hover_Throttle} shows the throttle for hovering condition and the corresponding attitude as well as current and voltage measurements are shown in Fig. \ref{Hover_Orient} and Fig. \ref{Hover_Vol_} respectively. Fig. \ref{Hover_Orient} shows the dynamic stability of the drone during rest and throttle. The pitching condition has also been tested for the drone, with the corresponding throttle, attitude and current-voltage plots shown in Fig.~\ref{pitchthrottle}, Fig. \ref{throttle} and Fig. \ref{pitchvc} respectively. The attitude plot shows the coupling between pitch and yaw angle. The drone's power consumption is approx. $65~W$ for a throttle of about $11~\%$ during hover and approx. $88~W$ for a throttle of about $14~\%$ during pitching. A video of the experiment during the hovering condition can be seen by clicking: \href{https://drive.google.com/file/d/1b4AG0teMeg9Wy_YH93Td3ekoLq9vQxrk/view?usp=drive_link}{Experiment Video}

\section{Conclusion} \label{sec:conclusion}
This paper presents the design process for a hybrid tilt-rotor bicopter drone, along with the required experimental validation and performance analysis. The drone has been designed with in-depth stress analysis and an optimized design to achieve static as well as dynamic stability. This paper extensively examines the modular design of the bicopter, including key structural and aerodynamic aspects. The structural study confirms the stresses' enduring capabilities during various flight modes. The thrust and power analysis provides an assessment of the flight performance and endurance. Following the analysis, a prototype of the drone was built and experiments were performed to verify basic attitude stability and power consumption. In the future, suitable path-planning and control algorithms will be designed and implemented to ensure trajectory tracking during flight tests.
\bibliographystyle{IEEEtran}
\bibliography{bibieee}



\end{document}